\documentclass[10pt,twocolumn,letterpaper]{article}

\usepackage{wacv}
\usepackage{times}
\usepackage{epsfig}
\usepackage{graphicx}
\usepackage{amsmath}
\usepackage{amssymb}
\usepackage{booktabs}

\usepackage{url}
\usepackage{booktabs}
\usepackage{algorithm}
\usepackage{algorithmic}

\usepackage{comment}
\usepackage{subcaption}
\usepackage[accsupp]{axessibility}  

\usepackage{multirow}
\usepackage{xcolor}
\usepackage{parskip}
\def\wacvPaperID{7} 

\wacvfinalcopy 

\ifwacvfinal
\usepackage[pagebackref=true,breaklinks=true,colorlinks,bookmarks=false]{hyperref}
\else
\usepackage[pagebackref=true,breaklinks=true,colorlinks,bookmarks=false]{hyperref}
\fi

\definecolor{blue}{RGB}{101,102,246}
\definecolor{red}{RGB}{240,87,62}
\definecolor{green}{RGB}{0,205,152}
\definecolor{violet}{RGB}{173,90,246}
\definecolor{white}{RGB}{255,255,255}

\begin{document}

\title{The Impact of Racial Distribution in Training Data on Face Recognition Bias: A Closer Look}

\author{Manideep Kolla\\
HyperVerge Inc.\\
{\tt\small manideep@hyperverge.co}
\and
Aravinth Savadamuthu\\
HyperVerge Inc.\\
{\tt\small aravinth.muthu@hyperverge.co}
}

\maketitle
\thispagestyle{empty}

\begin{abstract}
   Face recognition algorithms, when used in the real world, can be very useful, but they can also be dangerous when biased toward certain demographics. So, it is essential to understand how these algorithms are trained and what factors affect their accuracy and fairness to build better ones. In this study, we shed some light on the effect of racial distribution in the training data on the performance of face recognition models. We conduct 16 different experiments with varying racial distributions of faces in the training data. We analyze these trained models using accuracy metrics, clustering metrics, UMAP projections, face quality, and decision thresholds. We show that a uniform distribution of races in the training datasets alone does not guarantee bias-free face recognition algorithms and how factors like face image quality play a crucial role. We also study the correlation between the clustering metrics and bias to understand whether clustering is a good indicator of bias. Finally, we introduce a metric called racial gradation to study the inter and intra race correlation in facial features and how they affect the learning ability of the face recognition models. With this study, we try to bring more understanding to an essential element of face recognition training, the data. A better understanding of the impact of training data on the bias of face recognition algorithms will aid in creating better datasets and, in turn, better face recognition systems.
  
\end{abstract}

\section{Introduction}
\label{sec:intro}

Racial bias in face recognition (FR) systems is a widely acknowledged problem. Multiple studies have shown the racial inequity of these systems and the social injustice it translates into. In January 2020, Detroit police arrested Robert Williams in front of his family. It soon turned out that Mr. Williams was innocent and had been wrongfully arrested because an FR system had wrongly matched him with a suspect's face, which started the devastating series of events that happened \cite{RobertWilliams, hill2020wrongfully}.

Studies in the past have shown that there have been multiple orders of magnitude of improvement in the performance of face recognition algorithms over the last two decades \cite{grother2010mbe}. Despite these improvements, demographic bias in these algorithms is still a critical problem and affects real people. A recent study from NIST \cite{grother2019face} analyzed hundreds of algorithms and documented the difference in accuracies across different races. The study has shown that the false match rates (FMR) vary between 10 to 100 times across different demographics, and the false match differentials are much higher than that of false non-match, which vary by a factor of at most 3. The study reports that the East African, West African, and East Asian cohorts have the highest false match rates, and the Eastern European cohort has the lowest. This study has also shown that these algorithms falsely identified Native Americans more often than people from other demographics from U.S. law enforcement images. One interesting finding is that many algorithms built by developers from China have shown different characteristics compared to other algorithms with lower false match rates on East Asian faces, indicating how the algorithms are built, and the sourcing of the training data plays an essential role in determining their performance.

The causes of bias in face recognition systems are multidimensional, with statistical, human, and systemic biases. Many modern face recognition systems rely on large-scale data, which are generally scrapped from the internet. This has led to availability as the prime criterion taking precedence over suitability to the task \cite{schwartz2022towards}.

American Civil Liberties Union said in a statement that ``One false match can lead to missed flights, lengthy interrogations, watch list placements, tense police encounters, false arrests or worse" \cite{singer2019many}. Facial recognition technology has a particularly high potential to cause harm when it targets children because it is less accurate at identifying children \cite{barrett2020ban, srinivas2019face}. To build robust algorithms that are fair and work for everyone, we need to curate more real-world datasets and examine the existing ones.

In recent years, numerous studies have empirically analyzed the bias in face recognition and face attribute analysis systems, with insights on what contributes to bias and ways to measure it \cite{krishnapriya2020issues, nagpal2019deep, bowyer2019face, robinson2020face, terhorst2020face, bar2009role, kortylewski2018empirically, karkkainen2021fairface}. Research has also shown promising results in mitigating bias in these systems \cite{gong2020jointly, wang2020mitigating, gong2021mitigating, li2021learning}.

Given the societal implications of biased face recognition algorithms, it becomes critical to study the datasets used for training these models and how they affect recognition performance. To bring more focus and resolve to bias in face recognition, we extensively study the effects of using different racial distributions in the training datasets on the face recognition models in terms of accuracy, clustering metrics, intra and inter race similarities, face quality, UMAP projections, and decision thresholds.

\section{Related Work}
\label{sec:related}
With the advancements in deep learning, the new age face recognition (FR) algorithms have come a long way reaching near human-level performance and performing better than humans in tasks like large-scale face search systems. Methods proposed in \cite{taigman2014deepface, sun2015deep, Parkhi15, schroff2015facenet, wen2016discriminative, deng2019arcface} use different variations of neural networks, mainly convolutional neural networks, to embed the input faces into d-dimensional feature vectors, which would serve as a low-dimensional representation of faces, encoding different features of the face. There are many ways to train a neural network to encode the faces into these vector representations in such a way that these feature vectors can be used to compare the similarity between faces using metrics like euclidean and cosine distance.

Taigman \textit{et al.} \cite{taigman2014deepface} and Sun \textit{et al.} \cite{sun2015deep} propose to learn the vector representations by training a convolutional network using a softmax-based cross-entropy loss with identities as classes. In \cite{Parkhi15} and \cite{schroff2015facenet}, the vector representations of the faces are learned by training a convolutional network to minimize the triplet loss between an anchor, positive (mated), and a negative (non-mate) triplet of faces. Later, angular margin-based loss functions were proposed, which proved to be far more discriminative and accurately learn the face embeddings than the existing triplet-based and softmax-based losses without angular margin \cite{wen2016discriminative, deng2019arcface}.

Although the current FR algorithms are very accurate and cross human-level performance, many recent studies have pointed out the biases these FR algorithms learn and how they affect people from different demographics.

Krishnapriya \textit{et al.} \cite{krishnapriya2020issues} have shown that the false match rate (FMR) is higher in the African-American cohort, and the false non-match rate (FNMR) is higher in the Caucasian cohort at a fixed decision threshold. Apart from this, they have not found any clear evidence for the presumption that darker skin tone causes a higher false match rate. Nagpal \textit{et al.} \cite{nagpal2019deep} show that upon limited exposure to other races, face recognition algorithms mimic the human inclination of own-race bias. They also show that the networks trained on faces from a specific race encode the race-specific regions of interest. Robinson \textit{et al.} \cite{robinson2020face} show a notable boost in the overall performance of face recognition algorithms by learning subgroup-specific thresholds instead of using a global threshold. The experiments in \cite{terhorst2020face} shed light on face quality estimation and its effect on face recognition accuracy across demographics. Bar-Haim \textit{et al.} \cite{bar2009role} show that differences in morphological features have a more significant role than skin color in causing bias in face recognition in humans. A similar skin tone between a pair of images increases the likelihood of false matches, but a darker skin tone in itself is not responsible for the observation. Kortylewski \textit{et al.} \cite{kortylewski2018empirically} study the effect of dataset biases in the form of features like lighting and pose using controllable synthetically generated faces and show that these biases significantly impact the generalization of the FR algorithms.

Recently, several methods have been proposed to mitigate demographic bias in face recognition algorithms. Although the research on debiasing deep learning based FR algorithms is still nascent, these methods have shown great promise. Gong \textit{et al.} \cite{gong2020jointly} propose an adversarial training setup that extracts disentangled feature representations to address bias in face recognition. Wang \textit{et al.} \cite{wang2020mitigating} propose a reinforcement learning based race balance network to select appropriate margins for use in the large margin losses \cite{wen2016discriminative, deng2019arcface} for non-Caucasians to learn balanced performance for different races and published BUPT-GlobalFace and BUPT-Balancedface datasets to facilitate studies into bias in face recognition algorithms. In \cite{gong2021mitigating}, a training methodology with adaptive convolution kernels and attention mechanisms is proposed that adapt based on the demographic attributes of the faces to mitigate bias. Li \textit{et al.} \cite{li2021learning} formulate debiasing as a signal-denoising problem and propose a progressive cross-transformer architecture to denoise the identity-unrelated components induced by race from the identity-related components for fair face recognition.

Apart from algorithmic and architectural factors, one of the main reasons for biased face recognition algorithms is the presence of a non-uniform distribution of demographic classes, like racial distribution in the training datasets, which leads to unfair performance across different racial groups. In this study, to understand the effect of racial distribution in training datasets on the bias, we 1) train a face recognition model with a fixed architecture on different training datasets with different combinations of races, 2) compute the accuracy and clustering metrics of these trained models across all races to infer their bias, 3) study the correlation between bias and the clustering of faces based on race, 4) study the effect of face quality on the bias, 5) analyze the intra and inter race similarities, 6) visualize the clustering of faces from different races using a dimensionality reduction technique, and 7) study the difference in the decision thresholds of these trained models for different races and their correlation to bias.

\section{Methodology}
\label{sec:methodology}

\subsection{Network Architecture and Loss Functions}
\label{ssec:network}

We use the ResNet-50 network \cite{he2016deep, deng2019arcface} as the backbone for our experiments. The backbone consists of 43.6 million parameters. The output of the backbone network is a 512-dimensional (512-D) vector which serves as the embedding vector of an input face. The backbone is followed by a classification layer that outputs classification logits. We use Additive Angular Margin Loss (ArcFace) \cite{deng2019arcface} to modify the logits and softmax Cross-Entropy loss to calculate the final loss from the modified logits. 

\subsection{Training Data Preparation}
\label{ssec:train_setting}
Our goal is to understand the contribution of the faces from four available races in the BUPT-BalancedFace dataset \cite{wang2020mitigating} during training on bias in face recognition models. To this extent, we prepare 15 distinct training datasets from the BUPT-BalancedFace dataset by choosing combinations of all faces from one race at a time, two races at a time, three races at a time, and all four races at a time, as detailed in Tab. \ref{tab:accuracies}. We also train the model on the MS1MV3 dataset \cite{guo2016ms, deng2019lightweight}, which consists of highly imbalanced data with respect to its racial distribution.

\begin{table*}[t]
\small
\centering
\renewcommand{\arraystretch}{1.2}
\begin{tabular}{lccccccccc}
\toprule
\multirow{2}{*}{Training Data} & \multicolumn{3}{c}{Clustering Metrics $\uparrow$} & \multicolumn{6}{c}{Accuracy Metrics (in \%)}\\
\cmidrule(lr){2-4} \cmidrule(lr){5-10}
  & CH-All & CH-T & CH-NT & African & Asian & Cauc. & Indian & All & STD\\
    
\midrule
African+Asian+Caucasian+Indian     & 293.5 &293.5 &- & 94.85 & 94.37 & 96.97 & 95.48 & 95.13 & 1.13 \\

MS1MV3    & 199.5 &199.5 &- & 96.75 & 96.42 & 99.02 & 97.32 & 97.07 & 1.16 \\

\midrule
African+Asian+Caucasian  & 329.1 &121.1 &- & 94.23 &94.65 & 96.55 & 92.05 & 93.42 & 1.85 \\

African+Asian+Indian   &511.1 &365.9 &- &93.57 &93.92 &92.08 &94.72 &92.88 & 1.10 \\

African+Caucasian+Indian &796.0 &348.4 &- &93.75 &85.60 &96.32 &94.45 &90.39 & 4.74 \\

Asian+Caucasian+Indian & 863.0 &218.3 &- & 83.05 & 93.30 & 96.03 & 94.23 & 86.56 & 5.85 \\

\midrule

African+Asian  &567.2 &158.2 &823.8 &92.50 &92.52 &90.87 &89.38 &90.05 & 1.50 \\

African+Caucasian  &867.0 &145.8 &1389.4 &92.78 &82.48 &95.35 &90.37 &87.95 & 5.56\\

African+Indian &922.8 &610.9 &1263.3 &92.25 &83.38 &90.43 &92.93 &88.57 & 4.37\\

Asian+Caucasian &989.4 &25.4 &1436.3 &79.68 &93.20 &95.17 &88.57 &84.40 & 6.89\\

Asian+Indian &1065.4 &70.8 &2167.2 &79.77 &92.17 &88.27 &92.48 &84.34 & 5.92 \\

Caucasian+Indian  &1250.0 &317.5 &2323.3 &81.28 &84.20 &94.67 &92.97 &84.73 & 6.54 \\

\midrule

African &1042.1 &- &1240.8 &89.83 &78.75 &86.77 &86.15 &84.31 & 4.70\\

Asian  &1377.6 &- &1513.5 &71.32 &89.23 &81.22 &79.73 &76.92 & 7.34\\

Caucasian &1332.7 &- &1474.4 &74.50 &77.73 &92.38 &84.33 &79.38 & 7.91\\

Indian  &1418.3 &- &1850.2 &73.88 &78.42 &83.92 &88.40 &79.47 & 6.34\\

\bottomrule
\end{tabular}
\caption{This table depicts the 16 experiments conducted with different racial distributions in the training data along with the Calinski-Harabasz (CH) index, accuracies, and standard deviation of accuracies across the four racial cohorts available. Here, CH-All, CH-T, and CH-NT indicate the CH index calculated on the RFW test set with faces from all four races, from the races that are part of the training, and from the races that are not a part of the training, respectively. CH index is not calculated when only faces from one race are present. $\uparrow$ indicates that the higher the CH index, the better the clustering.}
\label{tab:accuracies}
\end{table*}

\section{Experimental Settings}
\label{sec:exps}

\subsection{Implementation Details}
\label{ssec:impl}
\noindent \textbf{Datasets:} For training, we mainly use two datasets. First, the BUPT-BalancedFace dataset \cite{wang2020mitigating} contains face images with both identity and race labels. The dataset contains faces of people from four ethnic demographics: African, Asian, Caucasian, and Indian, with 7000 identities in each, with around 300,000 unique images in each and 1.25 Million images in total. Second, we use the MS1MV3 dataset \cite{guo2016ms, deng2019lightweight} as a general-purpose large-scale dataset for training. MS1MV3 contains around 5 Million images with approximately 91,000 identities. MS1MV3 is highly imbalanced with respect to its racial distribution and comprises 76.3\% Caucasian, 14.5\% African, 6.6\% Asian, and 2.6\% Indian \cite{wang2019racial}. For testing, we use the Racial Faces in-the-Wild (RFW) dataset \cite{wang2019racial} because it contains a uniform 6000 mated and 6000 non-mated pre-defined face pairs from each of the four ethnic demographics: African, Asian, Caucasian, and Indian. For both training and testing datasets, the faces are cropped and aligned using the RetinaFace face detector \cite{deng2020retinaface} to produce face crops of size $112 \times 112$.

\noindent \textbf{Training Settings:} We use Stochastic gradient descent (SGD) with an initial learning rate of 0.1, a momentum of 0.9, and a weight decay of $5 \times 10^{-4}$. We train all the experiments with a batch size of 512 for 35 epochs with a learning rate scheduler that decreases the learning rate tenfold at the 18th, 25th, 30th, and 33rd epochs.

\subsection{Evaluation Metrics and Protocols}
\label{ssec:evalprotocol}
\noindent \textbf{Accuracy metrics:} We use verification accuracy as the performance metric similar to \cite{gong2020jointly, gong2021mitigating, wang2020mitigating}. We use the pre-defined mated and non-mated pairs from the RFW dataset to calculate the accuracy of the trained models on all four races separately. We use 10-fold cross-validation to produce ten decision thresholds at which we attain the highest accuracy for each fold, and the final accuracies are calculated using the average of the 10-folds. Tab. \ref{tab:accuracies} contains the accuracy metrics on different races in the RFW dataset. We also calculate accuracy on the combined RFW dataset with all races combined. This is indicated by ``All" in Tab. \ref{tab:accuracies}. We use standard deviation between the accuracies across the four races to measure the magnitude of accuracy differentials between races across the models, which we use as one of the metrics to gauge bias \cite{gong2020jointly, gong2021mitigating, wang2020mitigating}.

\noindent \textbf{Clustering metrics:} We also compute the Calinski-Harabasz (CH) index \cite{calinski1974dendrite} to understand the clustering ability and, in turn, the discriminative power of each model across races. CH index helps us understand the bias in face recognition models by gauging inter-race and intra-race cluster distance \cite{gluge2020not}. The intuition is to quantify how well the faces are clustered with respect to a specific characteristic like race. A good clustering of faces with respect to a specific characteristic like race suggests that the model has learned race-specific features apart from the identity-specific ones, potentially making the model susceptible to discrimination against people from different races. A higher value of the CH index indicates good clustering and, in turn, more bias. We computed the CH index in three scenarios for each model in Tab. \ref{tab:accuracies}. First, the index is computed using faces from all four races of the RFW dataset, followed by using only the races present during the training for a particular experiment, and last, the races that are not a part of the training. This will help us understand the clustering power of the models on the faces of ethnicities that are part of the training and those that are not. We discuss this in more detail in Sec. \ref{ssec:cluster_disc} and study whether clustering metrics like the CH index are an indicator of bias in FR algorithms. The clustering metrics are computed and detailed in Tab. \ref{tab:accuracies}.

\noindent \textbf{Racial Gradation:} We introduce the Racial Gradation metric to study the intra and inter race correlation in facial features across races. Gradation in facial features can be defined as the extent of similarity between faces from different races. To quantitatively measure the gradation, we calculate the intra and inter race average non-mated pair cosine distance across all the races by randomly sampling 100,000 non-mated pairs from the RFW test set for each intra and inter race setting. For a particular race A, if the inter-race average non-mated cosine distance is the lowest with a certain other race B, intuitively, the face recognition algorithm should be able to learn to recognize certain features from one race when it is trained only with the faces from the race with which it has lowest average distance. These intra and inter race average non-mated pair cosine distances across different races are depicted in Fig. \ref{fig:neg_distances}.

\noindent \textbf{UMAP projections:} We use the Uniform Manifold Approximation and Projection (UMAP) \cite{mcinnes2018umap} to visualize how the faces from different races are clustered by different models trained with varying racial distribution by projecting higher dimensional face embedding vectors to lower dimensions. To do this, we extract the 512-dimensional embedding vectors of the faces from the RFW test set and use the UMAP algorithm to project these embeddings onto a 2-dimensional plane. This particularly helps us analyze how the clustering of faces differs from model to model depending on the training data settings as depicted in Fig. \ref{fig:umaps}.

\noindent \textbf{Face quality:} To further understand whether face image quality plays a role in inducing bias, we calculate the face quality scores for training and test data using the face image quality assessment (FIQA) method proposed in \cite{ou2021sdd}. This gives a positive quality score for each face; the higher the score, the better the face quality. Although face quality is an important factor to study, we should also note that this FIQA model might be biased in itself due to the data it is trained on. For train and test sets, Fig. \ref{fig:quality_scores} contains the distribution of these quality scores, and Tab. \ref{tab:quality_scores} contains the median and mean face quality scores.

\noindent \textbf{Decision Thresholds:} Finally, graphs in Fig. \ref{fig:threshold_bar_graphs} show the cosine distance decision thresholds for each model at a false match rate (FMR) of 0.1\% for each race in the RFW test set.

\section{Results and Discussion}
\label{sec:discussion}

\subsection{On Accuracy}
\label{ssec: acc_disc}

We can observe an obvious pattern from Tab. \ref{tab:accuracies} that, on average, the standard deviation is the highest for the models trained on a single race and lowest in the models trained on all races. Within each data setting in Tab. \ref{tab:accuracies}, even when the settings concerning the racial distribution are similar, the standard deviations still vary a lot. In particular, the models trained with African faces in the training set have lower standard deviations, followed by Indian faces, and models trained with Caucasian faces have some of the highest standard deviations. This is true for all the data settings, from single-race training to training with three races. In particular, among the models trained on three races, the standard deviation is the lowest when trained with African and Indian faces, followed by the models trained with African and Asian faces. The models trained with Caucasian and without African faces have the highest standard deviation. This phenomenon can be attributed to two factors. First, the quality of the face images in training and test sets which we discuss in Sec. \ref{ssec:quality_disc}. Second, the gradation in facial features between different races is discussed in Sec. \ref{ssec:gradation_disc}.

The standard deviations between the model trained on the full BUPT-BalancedFace dataset and MS1MV3 are similar, even though the MS1MV3 is racially imbalanced. This is because MS1MV3 is a much larger dataset than BUPT-BalancedFace, and the model trained on the former inherently has lower error rates than when trained on the latter and, in turn, lower standard deviation. As observed here, the model trained on a smaller dataset, such as BUPT-BalancedFace face but with balanced racial distribution, has a standard deviation similar to that of the MS1MV3 model, even though the latter has lower error rates across all races. This indicates that the models trained on larger sets have lower absolute differences in accuracies between races, but this does not inherently mean that they are less biased.

\begin{figure}
  \centering
  \begin{subfigure}{0.49\linewidth}
    \includegraphics[width=1\textwidth]{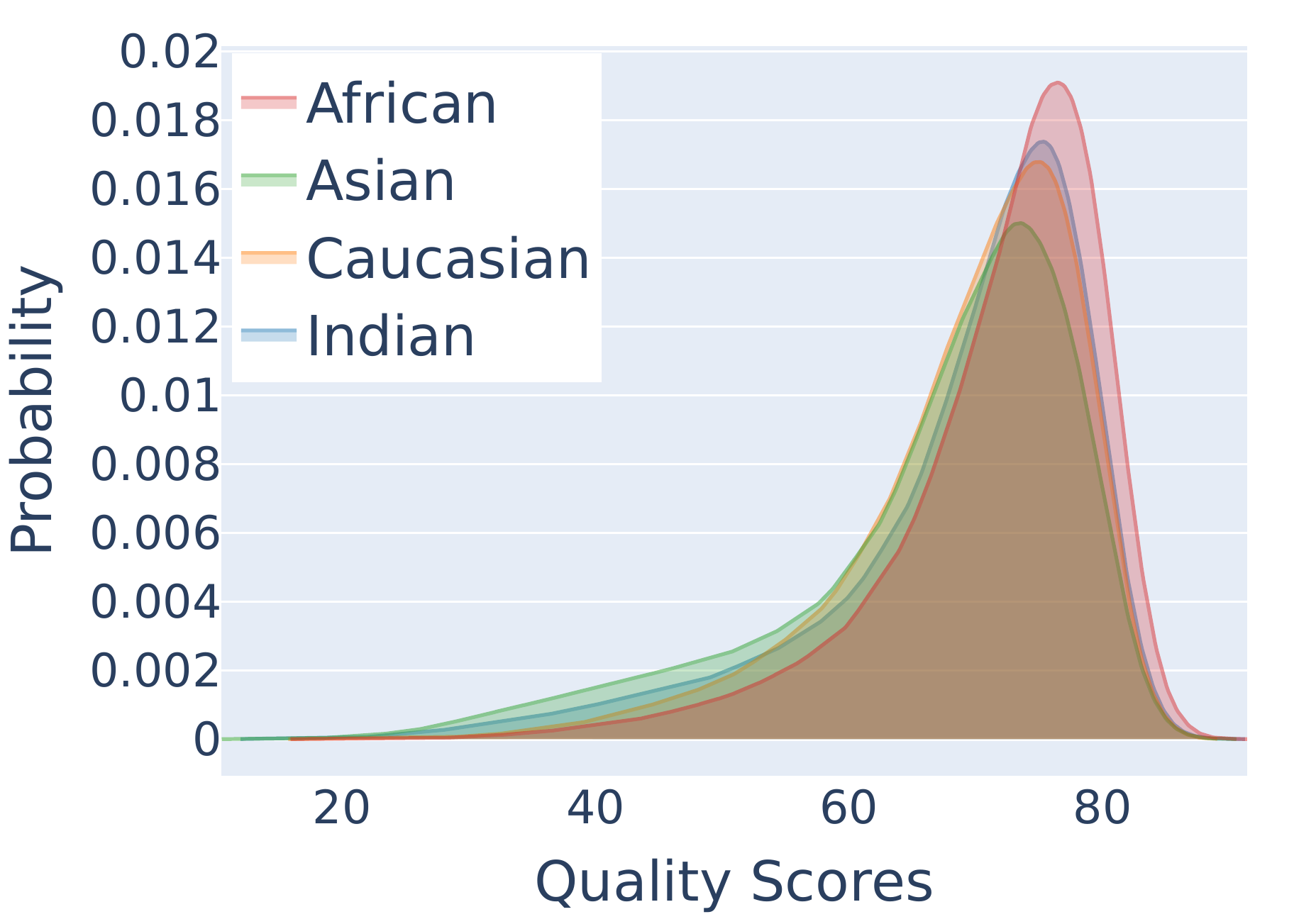}
    \caption{Training Data}
    \label{fig:train_quality_scores}
  \end{subfigure}
  \hfill
  \begin{subfigure}{0.49\linewidth}
    \includegraphics[width=1\textwidth]{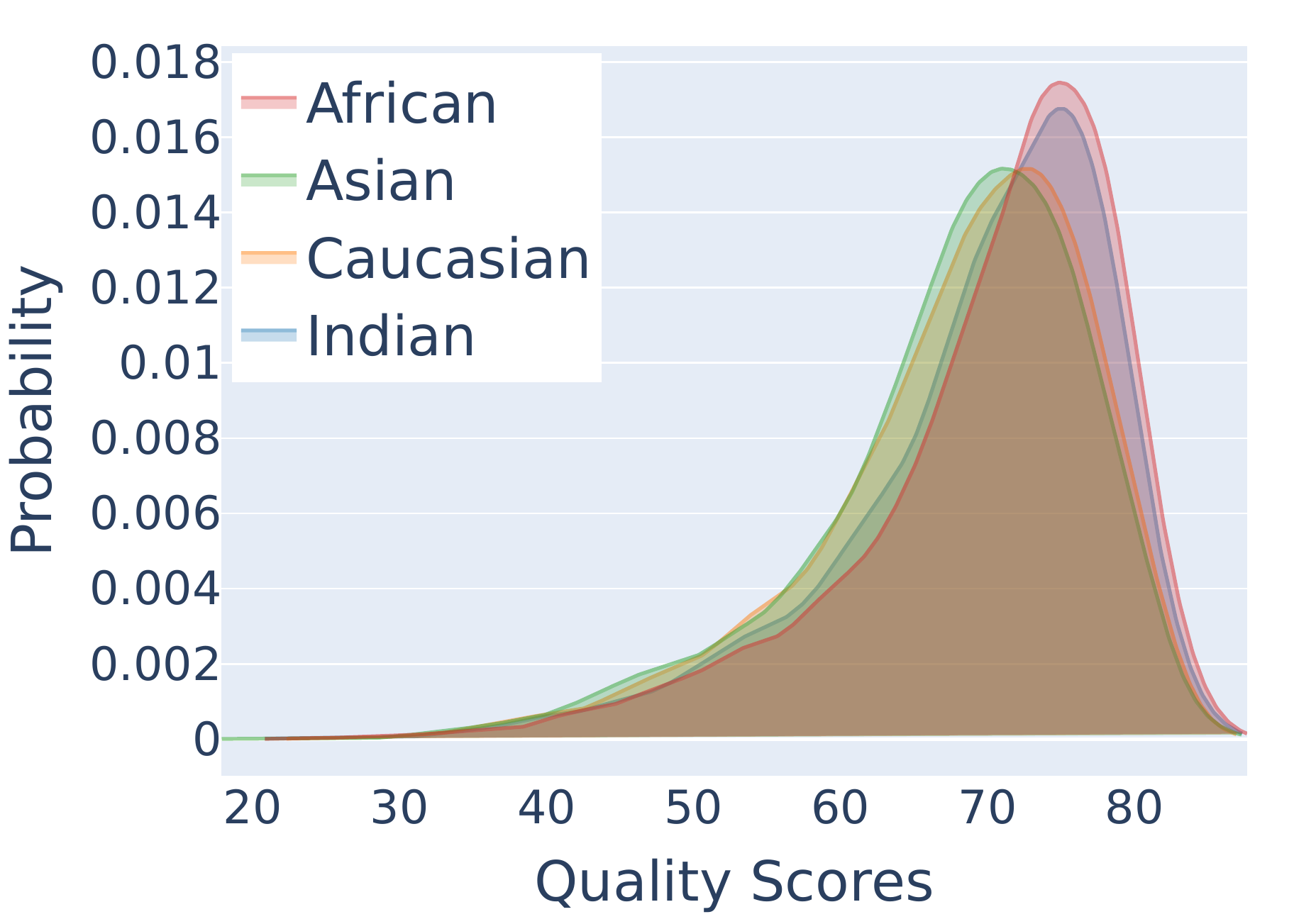}
    \caption{Test Data}
    \label{fig:test_quality_scores}
  \end{subfigure}
  \caption{Face quality scores distribution}
  \label{fig:quality_scores}
\end{figure}

\subsection{On Clustering Metrics}
\label{ssec:cluster_disc}
From Tab. \ref{tab:accuracies}, we can observe that the CH index is lower for in-train races and higher for out-of-training races. This indicates that the out-of-training races are getting clustered better than those in training, which contradicts the hypotheses made in Sec. \ref{ssec:evalprotocol}. We can observe from Tab. \ref{tab:accuracies} that the CH index and standard deviation do not have a monotonic relationship similar to the finding in \cite{gluge2020not}. In addition to this, to understand the extent of correlation between the CH index and standard deviation, we attempt to calculate Pearson's Correlation Coefficient \cite{benesty2009pearson} between these two metrics within each of the training data setting and across training data settings of Tab. \ref{tab:accuracies}. Pearson’s Correlation Coefficient is 0.921 across training data settings, indicating a very high positive correlation between the two metrics. Whereas within each training data setting, the correlation is inconsistent; however, has a high correlation for the most part. Within the four-race setting, Pearson’s Correlation is -1, indicating a perfect negative correlation as the number of settings for the four-race scenario is two, i.e, the sample size is two which are inversely correlated for calculating the Pearson’s Correlation. Pearson’s Correlation is 0.902 and 0.868 within the three and two race settings respectively indicating a high positive correlation between the two metrics. The positive correlation is the lowest in the one-race setting, with a Pearson’s Correlation of 0.775. These Pearson’s Correlation Coefficients indicate that except for the four race setting, there is a high correlation between the CH index and the standard deviation and particularly very high across training data settings from top to bottom of Tab. \ref{tab:accuracies} compared to within each setting. This finding is in contrast to the findings of \cite{gluge2020not} which reports no link between clustering metrics like CH index and standard deviation. Also, with an increasing CH index from four to single race settings, the overall accuracy on the full RFW test set decreases, indicating that models with better clustering ability might not be more accurate.

\begin{table}[t]
\small
\centering
\renewcommand{\arraystretch}{1.2}
\begin{tabular}{lcccc}
\toprule
\multirow{2}{*}{Race} & \multicolumn{2}{c}{Training Set} & \multicolumn{2}{c}{Test Set}\\
\cmidrule(lr){2-3} \cmidrule(lr){4-5}
  & Mean & Median & Mean & Median\\

\midrule
African   & 71.59 & 73.66 & 70.10 & 72.33\\

Asian     & 66.59 & 69.74 & 67.50 & 69.18\\

Caucasian & 69.26 & 71.26 & 67.92 & 69.62\\

Indian    & 68.73 & 71.69 & 69.33 & 71.36\\

\bottomrule
\end{tabular}
\caption{Mean and median face quality scores}
\label{tab:quality_scores}
\end{table}

\begin{figure*}
  \centering
  \begin{subfigure}{0.245\linewidth}
    \includegraphics[width=1\textwidth]{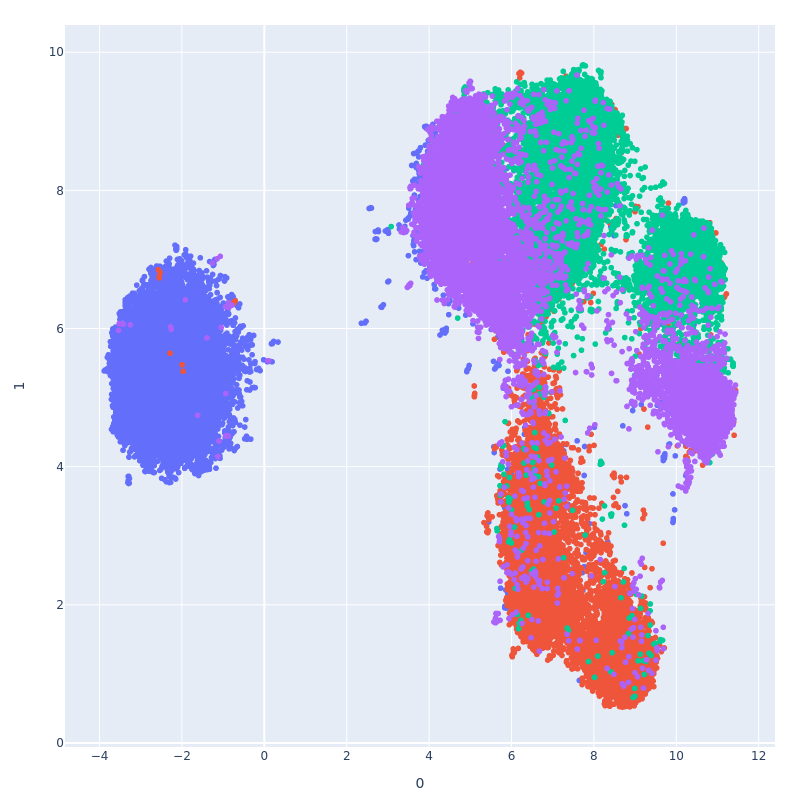}
    \caption{African}
    \label{fig:umap_African}
  \end{subfigure}
  \hfill
  \begin{subfigure}{0.245\linewidth}
    \includegraphics[width=1\textwidth]{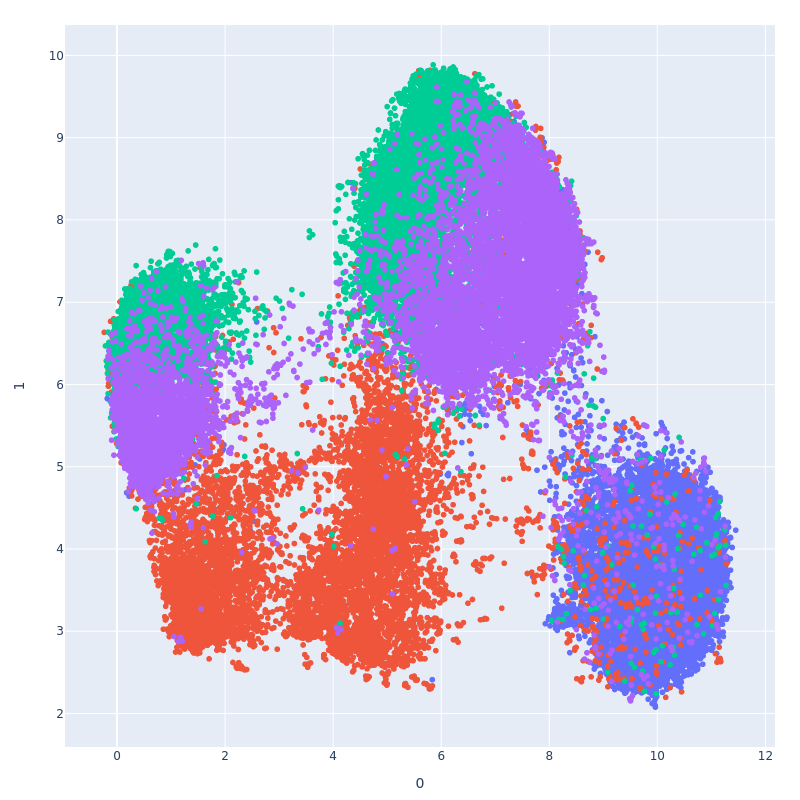}
    \caption{Asian}
    \label{fig:umap_Asian}
  \end{subfigure}
  \hfill
  \begin{subfigure}{0.245\linewidth}
    \includegraphics[width=1\textwidth]{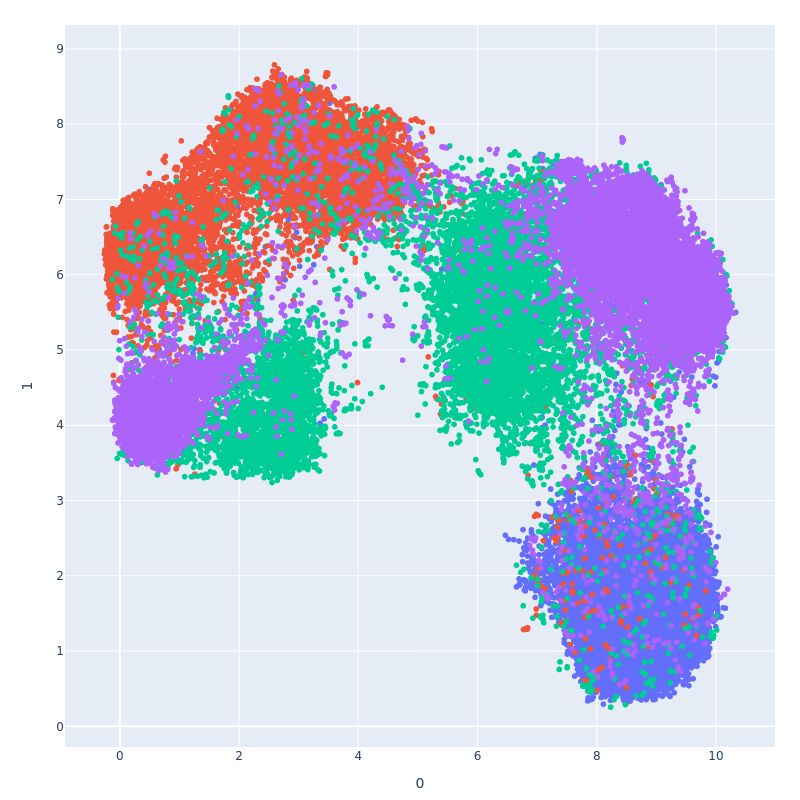}
    \caption{Caucasian}
    \label{fig:umap_Caucasian}
  \end{subfigure}
  \hfill
  \begin{subfigure}{0.245\linewidth}
    \includegraphics[width=1\textwidth]{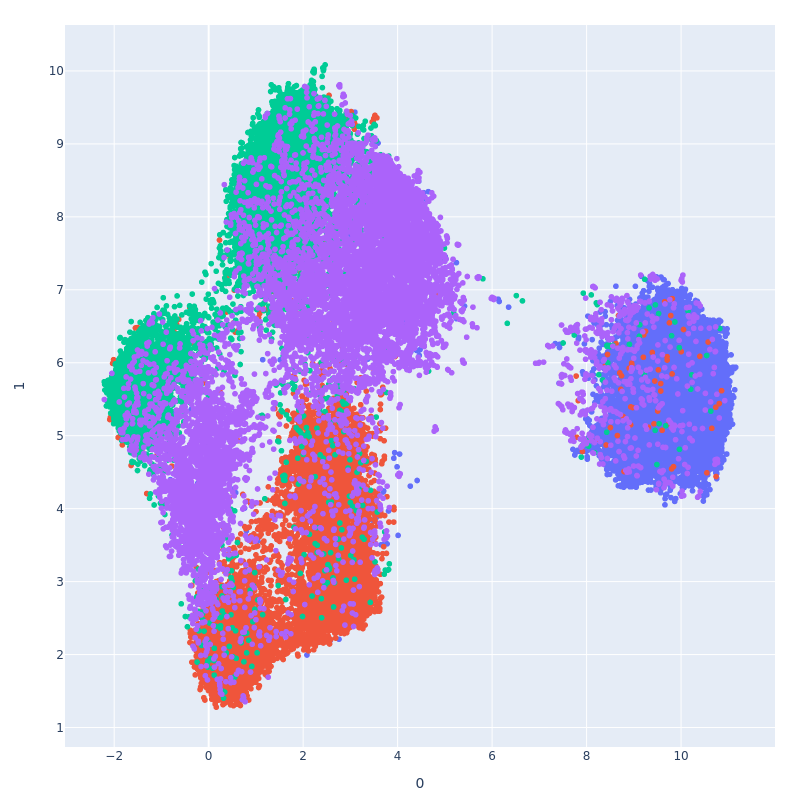}
    \caption{Indian}
    \label{fig:umap_Indian}
  \end{subfigure}
  \hfill
  \begin{subfigure}{0.245\linewidth}
    \includegraphics[width=1\textwidth]{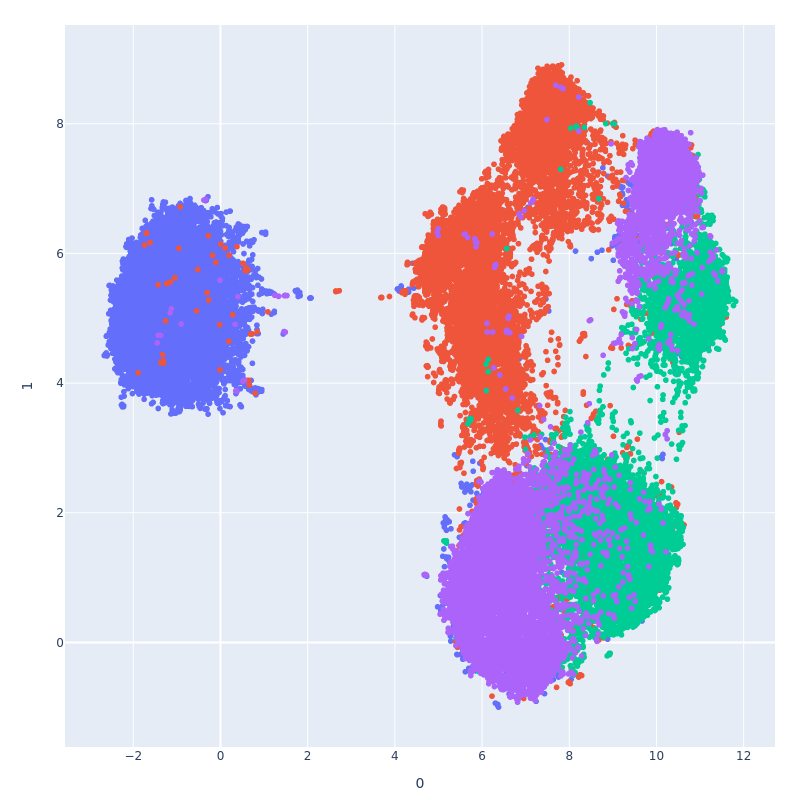}
    \caption{African+Asian}
    \label{fig:umap_African+Asian}
  \end{subfigure}
  \hfill
  \begin{subfigure}{0.245\linewidth}
    \includegraphics[width=1\textwidth]{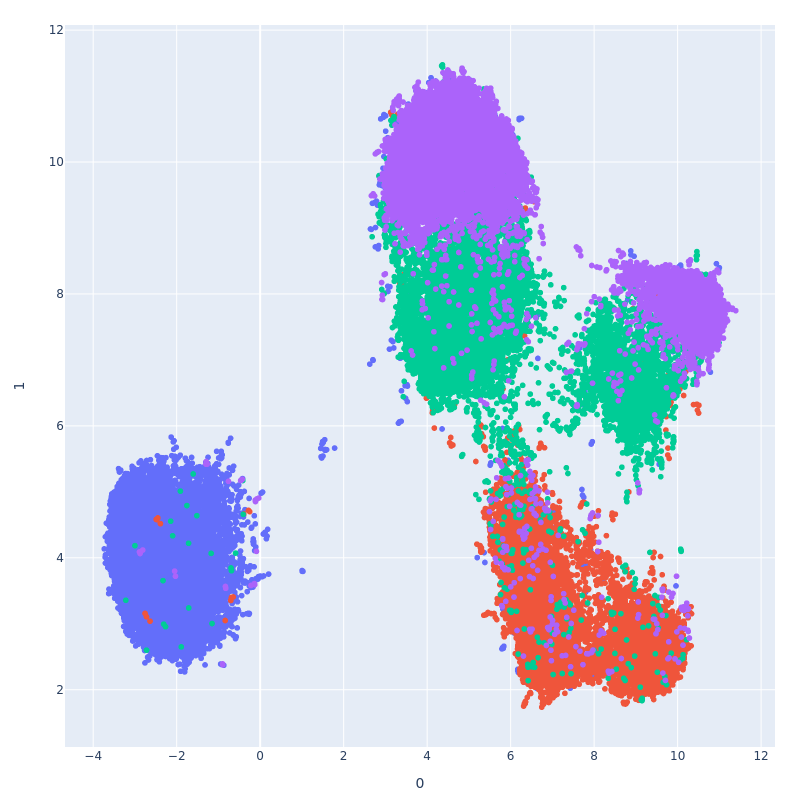}
    \caption{African+Caucasian}
    \label{fig:umap_African+Caucasian}
  \end{subfigure}
  \hfill
  \begin{subfigure}{0.245\linewidth}
    \includegraphics[width=1\textwidth]{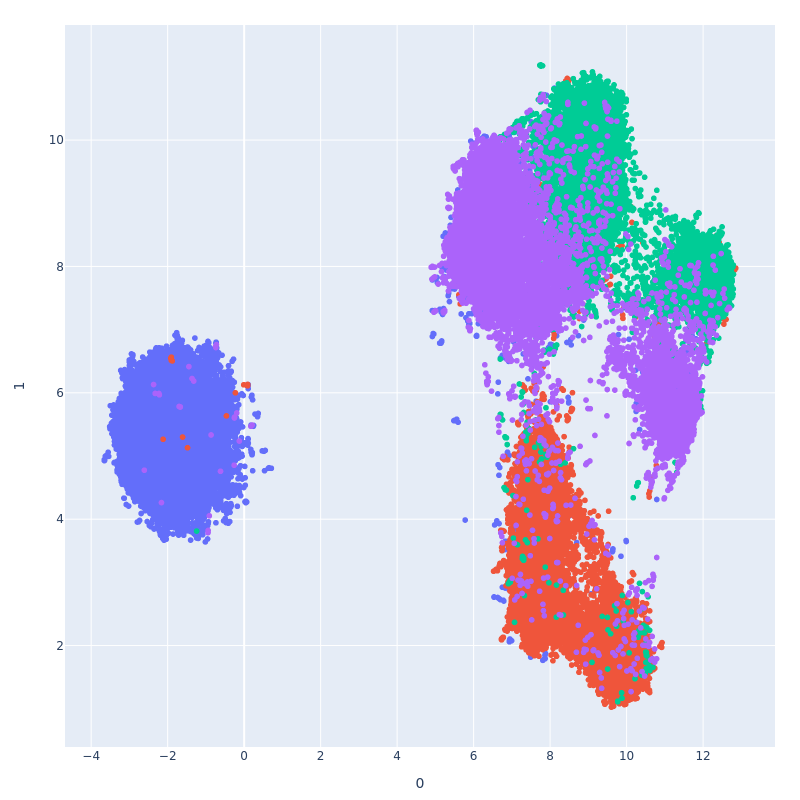}
    \caption{African+Indian}
    \label{fig:umap_African+Indian}
  \end{subfigure}
  \hfill
  \begin{subfigure}{0.245\linewidth}
    \includegraphics[width=1\textwidth]{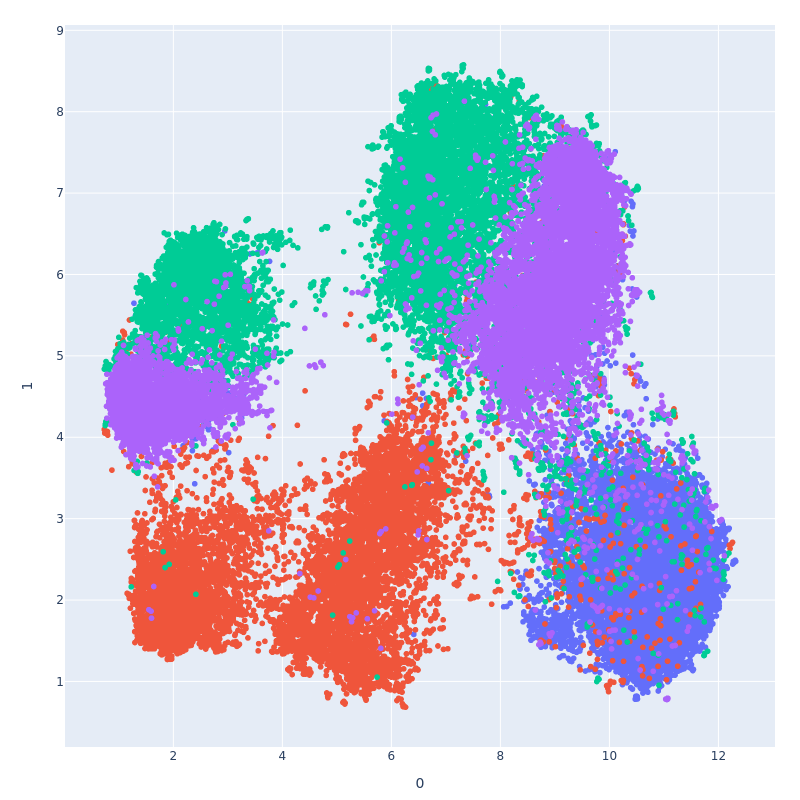}
    \caption{Asian+Caucasian}
    \label{fig:umap_Asian+Caucasian}
  \end{subfigure}
  \hfill
  \begin{subfigure}{0.245\linewidth}
    \includegraphics[width=1\textwidth]{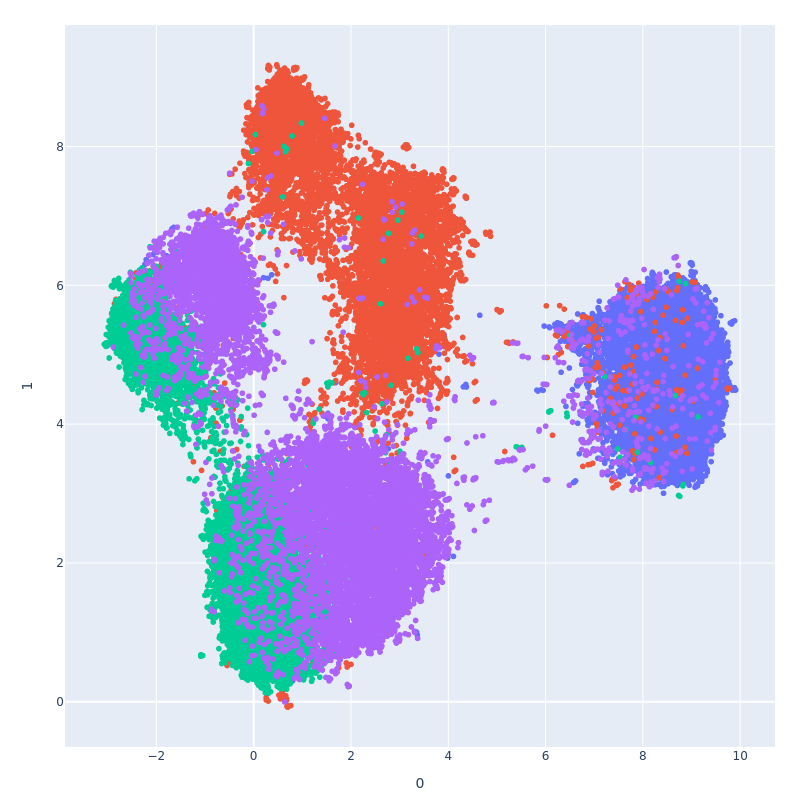}
    \caption{Asian+Indian}
    \label{fig:umap_Asian+Indian}
  \end{subfigure}
  \hfill
  \begin{subfigure}{0.245\linewidth}
    \includegraphics[width=1\textwidth]{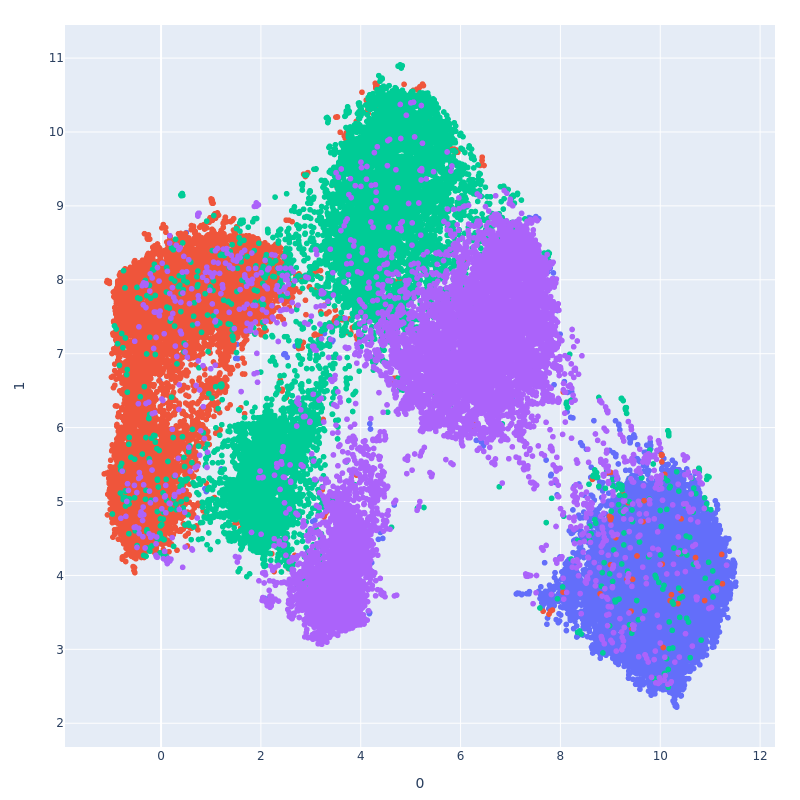}
    \caption{Caucasian+Indian}
    \label{fig:umap_Caucasian+Indian}
  \end{subfigure}
  \hfill
  \begin{subfigure}{0.245\linewidth}
    \includegraphics[width=1\textwidth]{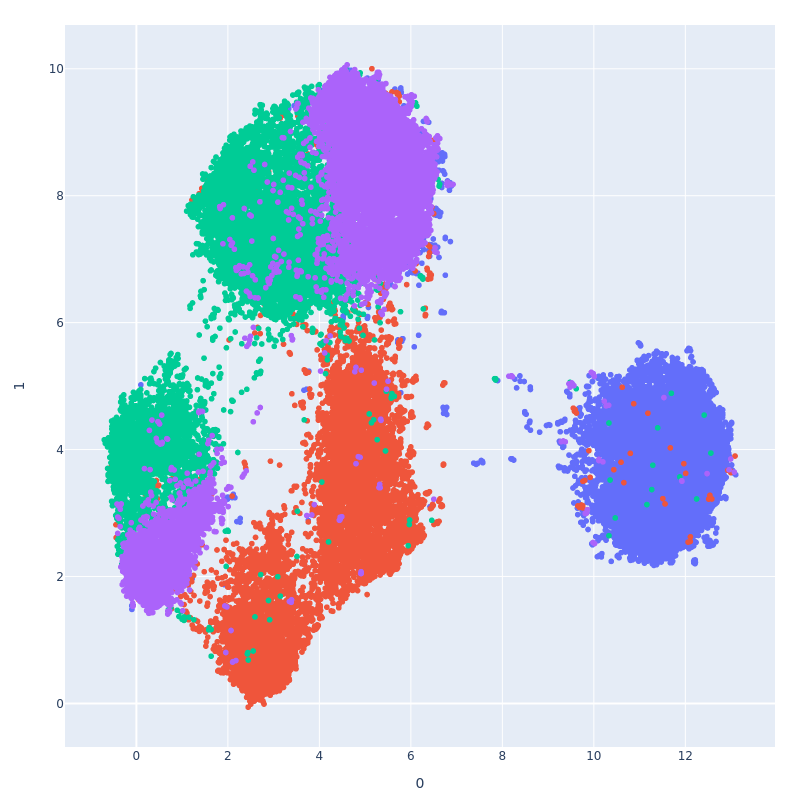}
    \caption{African+Asian+Caucasian}
    \label{fig:umap_African+Asian+Caucasian}
  \end{subfigure}
  \hfill
  \begin{subfigure}{0.245\linewidth}
    \includegraphics[width=1\textwidth]{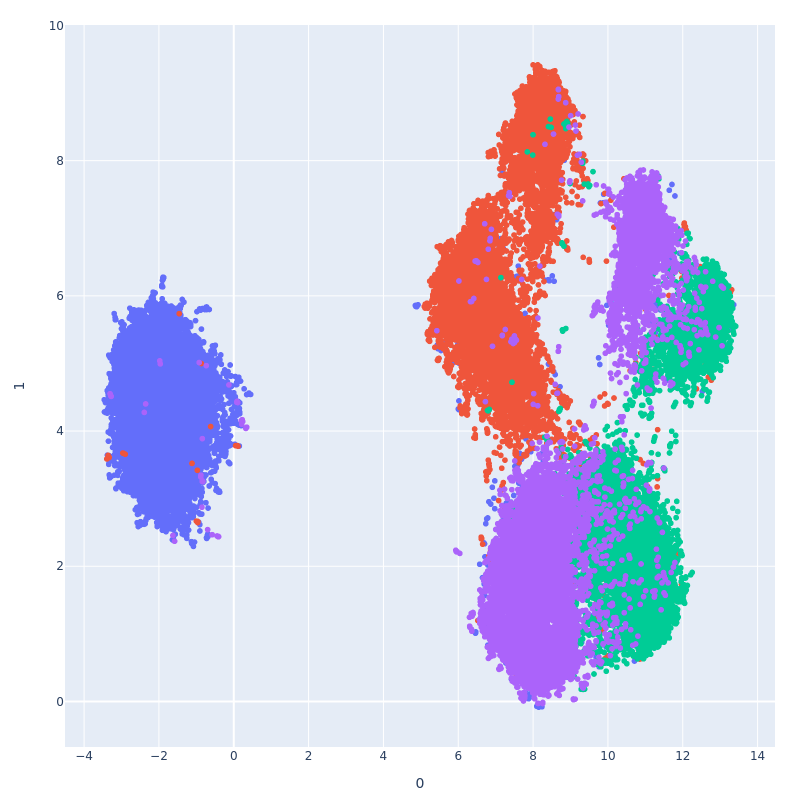}
    \caption{African+Asian+Indian}
    \label{fig:umap_African+Asian+Indian}
  \end{subfigure}
  \hfill
  \begin{subfigure}{0.245\linewidth}
    \includegraphics[width=1\textwidth]{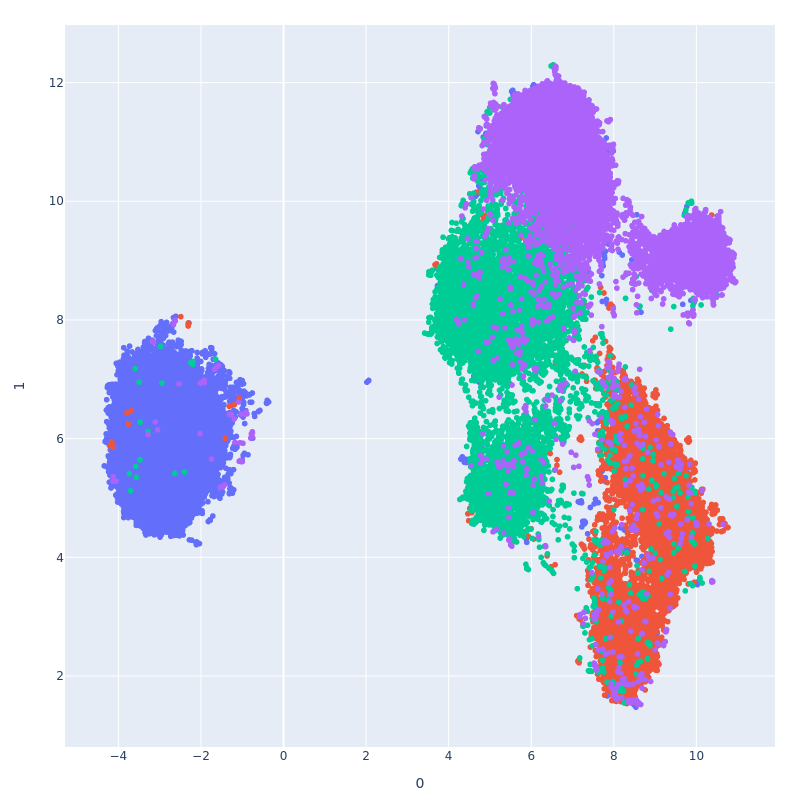}
    \caption{African+Caucasian+Indian}
    \label{fig:umap_African+Caucasian+Indian}
  \end{subfigure}
  \hfill
  \begin{subfigure}{0.245\linewidth}
    \includegraphics[width=1\textwidth]{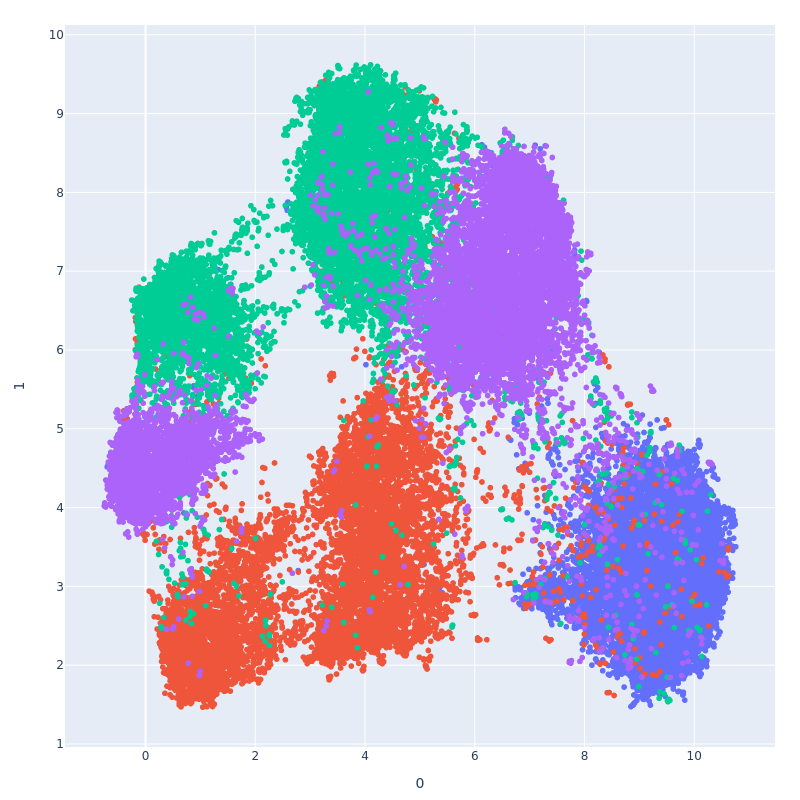}
    \caption{Asian+Caucasian+Indian}
    \label{fig:umap_Asian+Caucasian+Indian}
  \end{subfigure}
  \hfill
  \begin{subfigure}{0.245\linewidth}
    \includegraphics[width=1\textwidth]{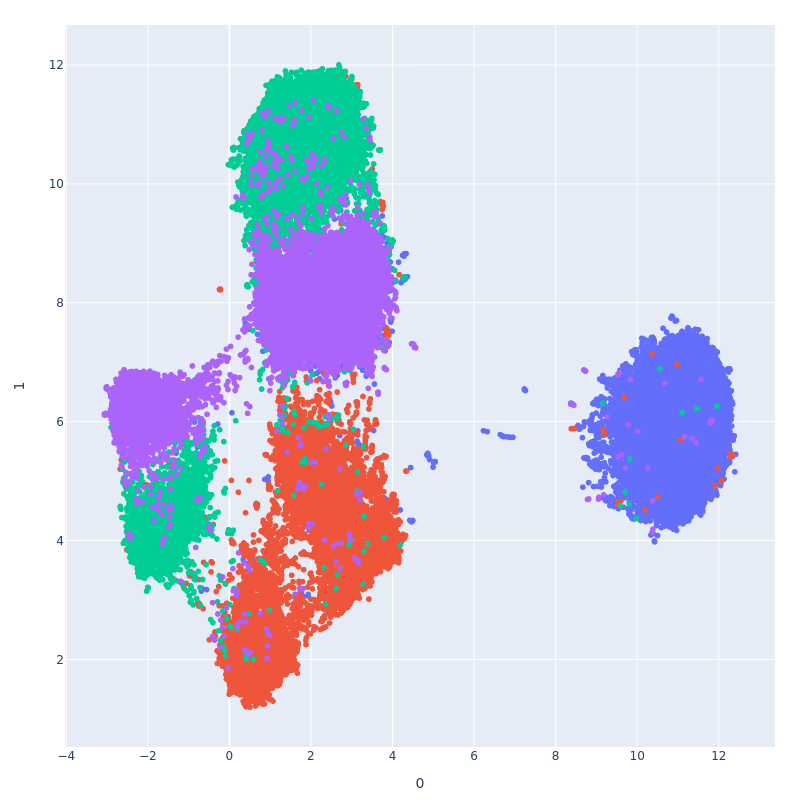}
    \caption{African+Asian+Cauc.+Indian}
    \label{fig:umap_African+Asian+Caucasian+Indian}
  \end{subfigure}
  \hfill
  \begin{subfigure}{0.245\linewidth}
    \includegraphics[width=1\textwidth]{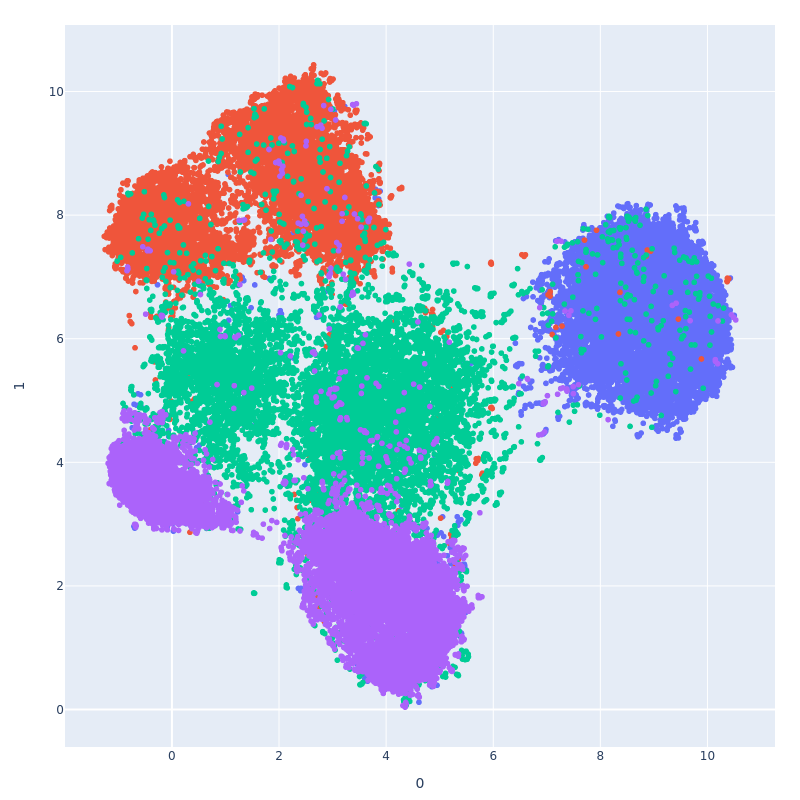}
    \caption{MS1MV3}
    \label{fig:umap_ms1m-wo-rfw}
  \end{subfigure}
  \hfill \textcolor{white}{.} \\ \textcolor{blue}{\textbf{$\bullet$ Blue}} indicate the African faces \hspace{1.2cm} \textcolor{red}{\textbf{$\bullet$ Red}} indicate the Asian faces \\ \textcolor{green}{\textbf{$\bullet$ Green}} indicate the Caucasian faces \hspace{1cm} \textcolor{violet}{\textbf{$\bullet$ Violet}} indicate the Indian faces.
  \caption{Two-dimensional UMAP projections of the face embeddings from the RFW test set computed using all the trained models.}
  \label{fig:umaps}
\end{figure*}

\subsection{On Dependence on Face Quality}
\label{ssec:quality_disc}
We can observe from Fig. \ref{fig:quality_scores} and Tab. \ref{tab:quality_scores} that African faces have the highest quality and Asian have the lowest quality in both train and test sets. This correlates with the observations made in Sec. \ref{ssec: acc_disc} that the models trained with African faces have significantly lower standard deviations than other races. This is because the models would be able to learn facial features better when the quality of the faces is better and will help in recognizing the faces from other races better than when the models are trained on faces of lower quality. We can particularly observe this in the single-race training setting, where the model trained with African faces also has the highest accuracies on other races after the models trained with the faces from those respective races.

\begin{figure}[t]
  \centering
  \includegraphics[width=\linewidth]{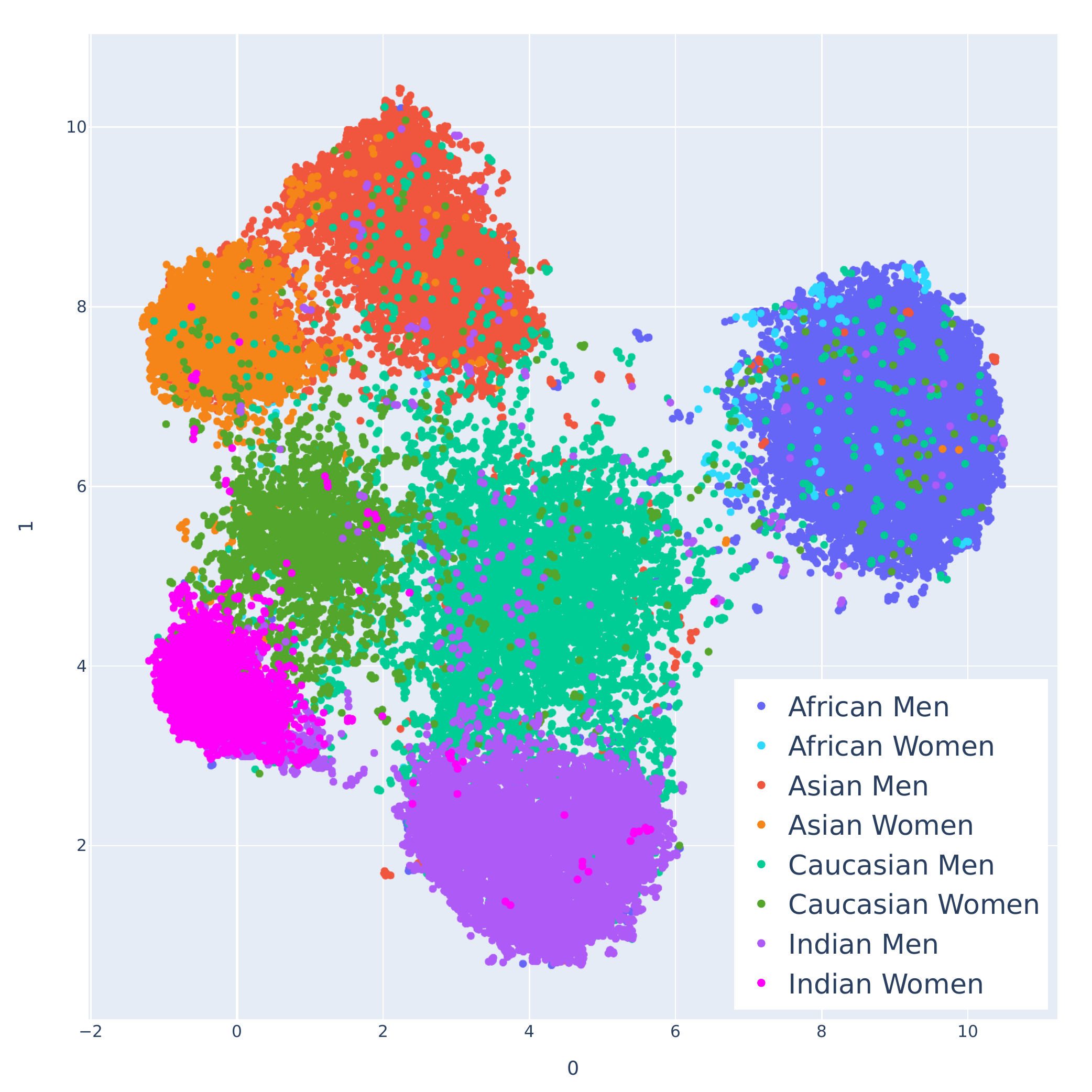}
  \caption{UMAP projections for the model trained on MS1MV3 data visualizing the gender sub-clusters.}
  \label{fig:umap_ms1mv3_with_genders}
\end{figure}

\subsection{On Gradation in Facial Features}
\label{ssec:gradation_disc}
An observation from Tab. \ref{tab:accuracies}, when the training data consists of African faces, the accuracy on the African test set is high and is comparable to the accuracy on the other three races when their respective data is used during training. However, when the African faces are not in the training set, the accuracy on the African test set is very low and is far lower than any other race in the test set in all training settings. The distinctly low accuracy on the African set in the absence of African faces from training results in a high standard deviation. Apart from face quality, we can attribute this to the fact that the African faces have much lower similarity to faces from other races, unlike other races, which have at least one neighboring race, as observed in Fig. \ref{fig:neg_distances}.

\subsection{On Correlation Between Gradation and Face Quality}
\label{ssec:gradation_quality}

\begin{figure}[t]
  \centering
  \includegraphics[width=\linewidth]{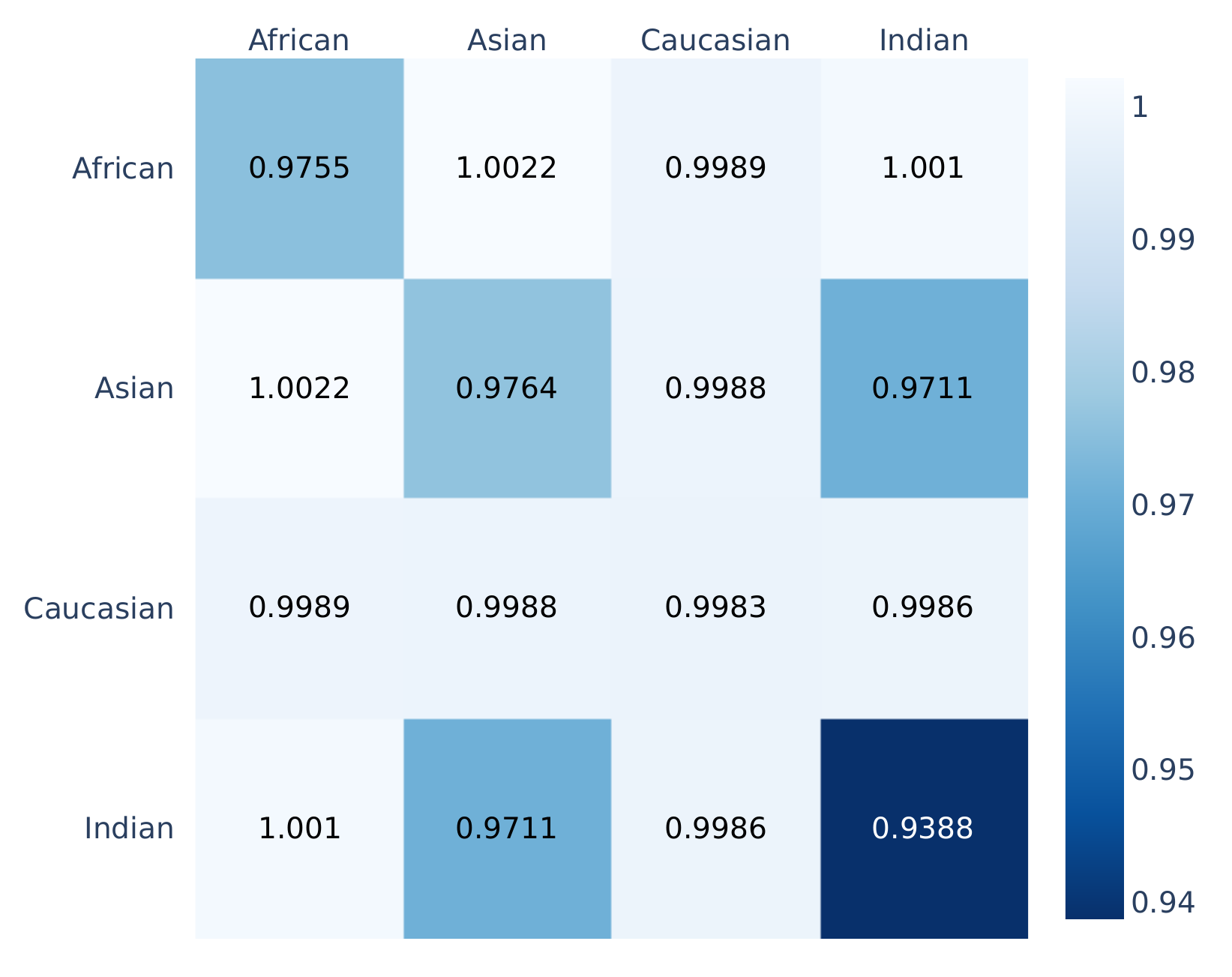}
  \caption{Gradation Matrix: Average cosine distances between non-mated face pairs across all races.}
  \label{fig:neg_distances}
\end{figure}

If two races are nearer, according to Fig. \ref{fig:neg_distances}, the accuracies on those races should be the highest when both the races are present together during training, compared to all other cases. Even if a particular race is absent in the training set, a nearer race from Fig. \ref{fig:neg_distances} would provide reasonable information to learn the former race. In two race settings of Tab. \ref{tab:accuracies}, when the training data consists of both Asian and Indian faces, the accuracy on both Asian and Indian test data is expected to be higher compared to all other settings, but this is not the observation from Tab. \ref{tab:accuracies}.

Indian accuracy in the Asian+Indian setting (92.48\%) is less than in the African+Indian setting (92.93\%). Similarly, Indian accuracy in the African+Asian setting (89.38\%) is less than in the African+Caucasian setting (90.37\%). Although this is an exception to the assumption made above from the gradation point of view, these exceptions can be explained using the face quality scores. Where in the first exception, African data has much better face quality than Asian, and in the second exception, Asian data has the lowest face quality of all. Similarly, Indian accuracy in the Asian+Indian setting (92.48\%) is marginally less than in the Caucasian+Indian setting (92.97\%) because the quality of Asian faces is lower than that of Caucasian faces. Though African face quality is higher than Caucasian face quality, Indian accuracy in the Caucasian+Indian setting (92.97\%) is marginally higher than that of the African+Indian setting (92.93\%) because Caucasian race is closer to Indian race compared to African as observed in Tab. \ref{fig:neg_distances}. Based on this, we conclude that face quality comes into play only when the quality difference is significant, like in the case of the first exception mentioned above, where African faces are of the highest quality, and Asian faces are of the lowest quality.

\begin{figure*}
  \centering
  \begin{subfigure}{0.49\linewidth}
    \includegraphics[width=1\textwidth]{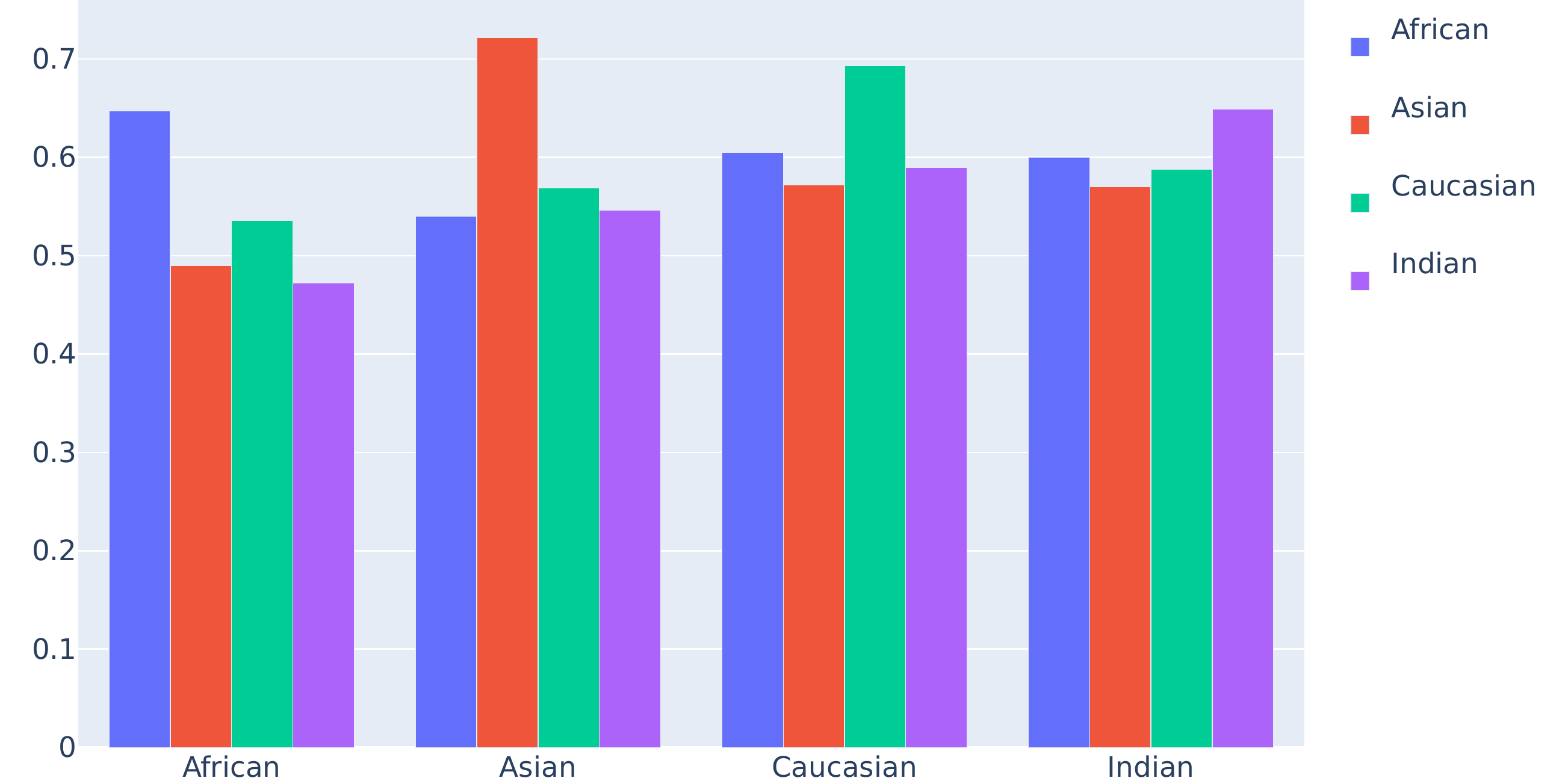}
    \caption{Models trained on data from single race.}
    \label{fig:one_race_thresholds}
  \end{subfigure}
  \hfill
  \begin{subfigure}{0.49\linewidth}
    \includegraphics[width=1\textwidth]{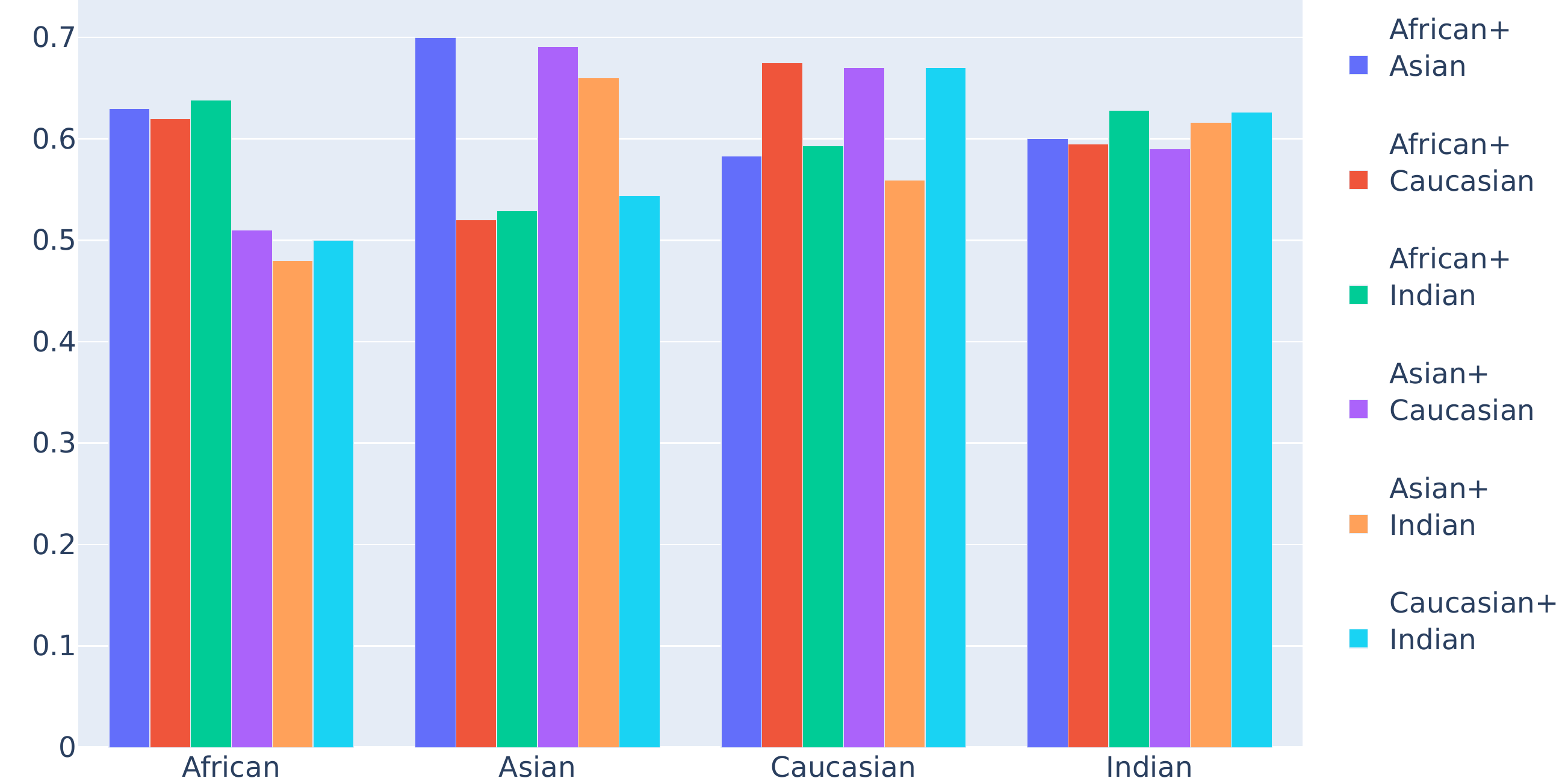}
    \caption{Models trained on data from two races.}
    \label{fig:two_race_thresholds}
  \end{subfigure}
  \hfill
  \begin{subfigure}{0.49\linewidth}
    \includegraphics[width=1\textwidth]{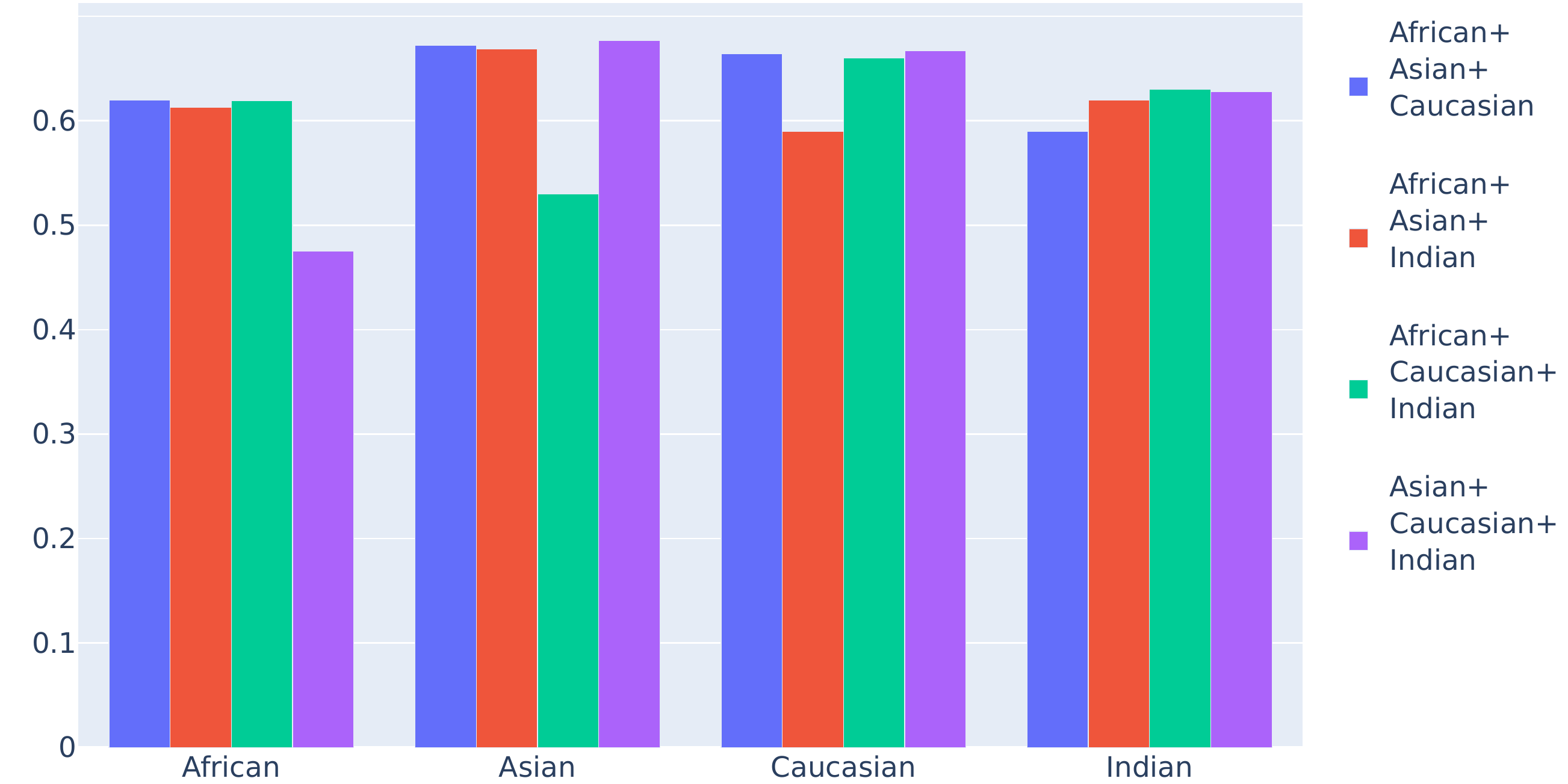}
    \caption{Models trained on data from three races.}
    \label{fig:three_race_thresholds}
  \end{subfigure}
  \hfill
  \begin{subfigure}{0.49\linewidth}
    \includegraphics[width=1\textwidth]{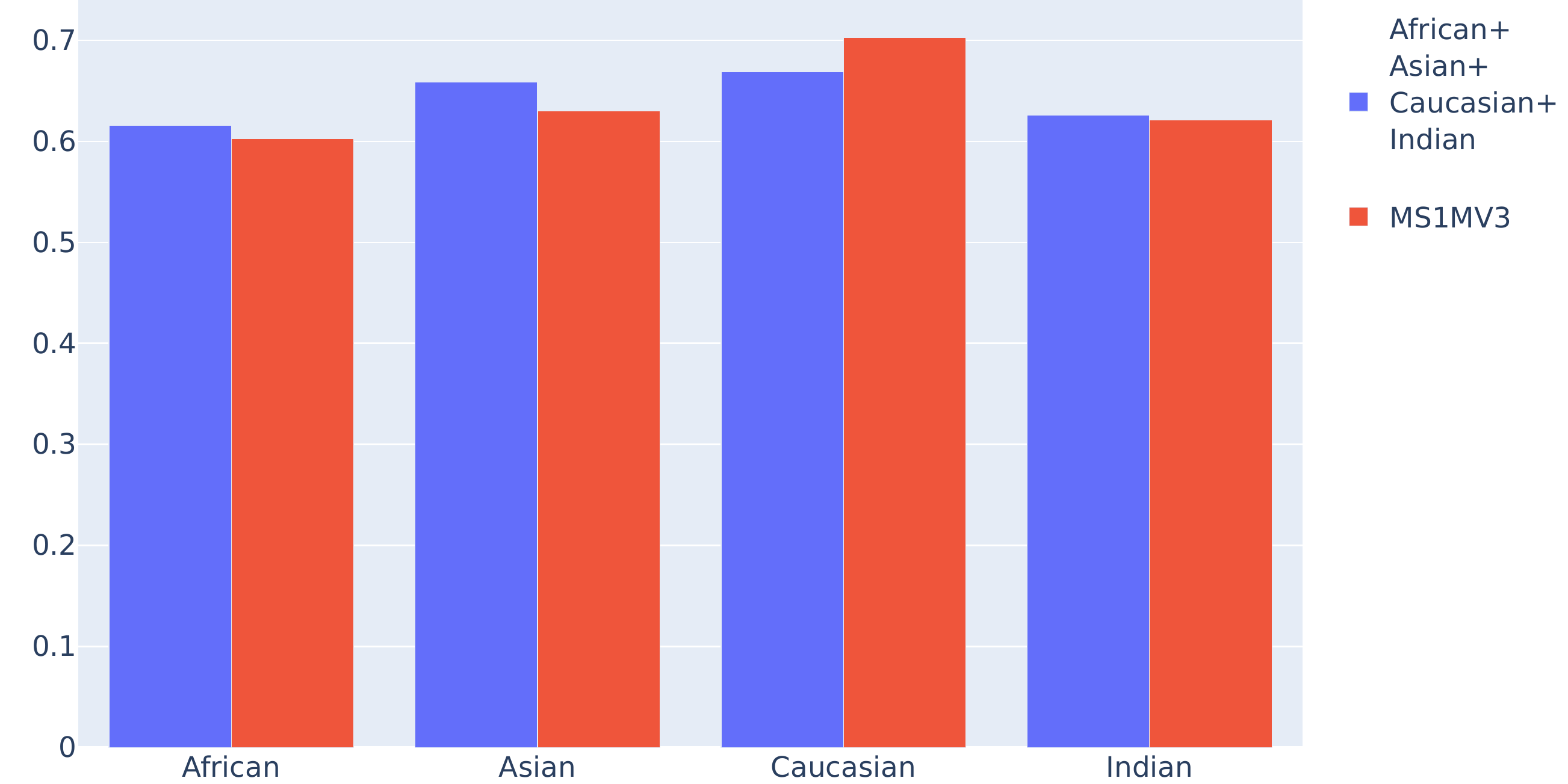}
    \caption{Models trained on data from all four races.}
    \label{fig:all_race_thresholds}
  \end{subfigure}
  \caption{Bar plots with the cosine distance decision thresholds of all the trained models on the four races of the RFW test set. The y-axis indicates the cosine distances, the x-axis indicates the races from the test set on which the thresholds are calculated, and each bar indicates a trained model identified by their training data racial distribution.}
  \label{fig:threshold_bar_graphs}
\end{figure*}

\subsection{On UMAP Projections}
\label{ssec:umap_disc}

With most models in Fig. \ref{fig:umaps}, African faces form a distinct and distant cluster. African faces are best clustered when African faces are present in the training set as expected and discussed in Sec. \ref{ssec:cluster_disc}. However, also when the training data has Indian faces, the African faces are clustered distantly, as seen in Fig. \ref{fig:umap_Indian}. This is because the Indian race is the farthest from the African race in Fig. \ref{fig:neg_distances}. When one of them is learned, the faces from these two races move apart as the features learned by the model from this race are very different from the other race. With most models in Fig. \ref{fig:umaps}, the Caucasian and Indian races are getting clustered together. Also, the faces of Asian, Caucasian, and Indian races are split into two sub-clusters each. To understand if gender plays a role in forming these sub-clusters, we predict the genders of all the faces in the RFW test set using \cite{serengil2021hyperextended}, which has a comparable gender prediction accuracy across races. Using this gender information, we plot the UMAP projections of faces in the RFW set with the model trained on the MS1MV3 dataset as shown in Fig. \ref{fig:umap_ms1mv3_with_genders}. This clearly indicates that these sub-clusters are due to gender, and the sub-clusters of the same gender are grouped together. African faces do not form visible sub-clusters because the representation of women in the African subset of the RFW test set is meager compared to other races to form a visible sub-cluster of women. The proportion of women in the African cohort is 1.62\%, whereas it is 31.69\%, 28.16\%, and 25.53\% for the Asian, Caucasian, and Indian cohorts of the RFW set.

\subsection{On Decision Thresholds}
\label{ssec:threshold_disc}
Fig. \ref{fig:threshold_bar_graphs} depict the decision thresholds for all the models trained with different data settings. From Fig. \ref{fig:one_race_thresholds}, we can observe that the decision threshold is the highest for the race using which the model is trained. This indicates that the model trained on faces from a certain race tends to be more confident in recognizing faces from that race and, in turn, has a higher decision threshold. This is true for all the training data settings. Similarly, in Fig. \ref{fig:all_race_thresholds}, the model trained on the MS1MV3 dataset has the highest decision threshold for the Caucasian cohort, whereas the model trained on all four races of the BUPT-BalancedFace dataset has comparable decision thresholds across all racial cohorts. This is because the MS1MV3 dataset has disproportionately more Caucasian faces, which is not the case for the BUPT-BalancedFace. This is also called own-race bias \cite{meissner2001thirty}.

\section{Conclusion}
\label{sec:conclusion}
In this work, we show how the racial distribution in the training data affects the racial bias of face recognition models. We show that the face recognition models perform worse on the faces belonging to the unseen ethnicities during training, but we find that having an equal representation of faces from different races does not guarantee bias-free algorithms. We empirically show that, apart from non-uniform racial distribution in training data, factors like face quality and gradation in facial features across races play a considerable role in inducing bias in these models. We also analyze whether the clustering of faces based on race is a good indicator of bias. Unlike previous studies, we find a very high correlation between the two across different training data settings and a high correlation within all training data settings except the four-race setting. Finally, we visualize the clustering of faces using UMAP projections, which uncovered the role of gender in clustering. In this study, we lay down various ideas on what to look for in the training data to understand bias in face recognition algorithms apart from just racial distribution. We hope our research will guide future work in understanding the bias in face recognition models through the lens of data and help curate more educated datasets.

{\small
\bibliographystyle{ieee_fullname}
\bibliography{egbib}
}

\end{document}